\documentclass[a4paper,fleqn]{cas-sc}

\usepackage[authoryear]{natbib}
\usepackage{graphicx}
\usepackage{adjustbox}
\usepackage{todonotes}
\usepackage[T1]{fontenc}
\usepackage{verbatimbox}
\usepackage{hyperref}
\hypersetup{
    colorlinks=true,
    linkcolor=blue,
    filecolor=magenta,      
    urlcolor=blue,
}

\usepackage{gensymb}
\usepackage{booktabs}
\usepackage{adjustbox}
\usepackage{lscape}
\usepackage{subcaption}
\usepackage{caption}
\usepackage{multirow}
\usepackage{algorithm}
\usepackage{algpseudocode}
\usepackage{booktabs}
\usepackage{amssymb}
\usepackage[utf8]{inputenc}
\usepackage{color}
\usepackage{amsmath}
\usepackage{lineno}
\usepackage{setspace}

\usepackage{booktabs}
\usepackage{multirow}

\def\tsc#1{\csdef{#1}{\textsc{\lowercase{#1}}\xspace}}
\tsc{WGM}
\tsc{QE}
\tsc{EP}
\tsc{PMS}
\tsc{BEC}
\tsc{DE}

\begin{document}
\let\WriteBookmarks\relax
\def\floatpagepagefraction{1}
\def\textpagefraction{.001}
\shorttitle{Two Scalable Approaches for Burned-Area Mapping}
\shortauthors{I. Mancila-Wulff et~al.}

\title [mode = title]{Two Scalable Approaches for Burned-Area Mapping Using U-Net and Landsat Imagery}

\author[1]{Ian 
Mancilla-Wulff}[orcid=0000-0002-1392-5569]
\ead{ian.mancilla@sansano.usm.cl}

\credit{Conceptualization; Methodology; Software; Formal analysis; Visualization; Writing – original draft}

\author[2,3]{Jaime Carrasco}[orcid=0000-0003-4123-4228]
\cormark[1]
\ead{jcarrascob@utem.cl}

\credit{Conceptualization; Methodology; Formal analysis; Writing, review, editing – original draft}

\author[3,4]{Cristobal Pais}[orcid=0000-0002-1392-5569]
\ead{cpaismz@berkeley.edu}

\credit{Conceptualization; Methodology}

\author[5,6]{Alejandro Miranda}[]
\ead{alejandro.miranda@ufrontera.cl}

\credit{Conceptualization; Data curation; Writing \& Review}

\author[1,3]{Andr\'es Weintraub}[]
\ead{aweintra@dii.uchile.cl}

\credit{Supervision; Writing \& Review}

\address[1]{Industrial Engineering Department, University of Chile,  Santiago, Chile}

\address[2]{Departamento de Industria, Facultad de Ingeniería, Universidad Tecnol\'ogica Metropolitana, Santiago, Chile}

\address[3]{Complex Engineering System Institute - ISCI, Santiago, Chile}

\address[4]{IEOR Department, University of California Berkeley, Berkeley, USA}

\address[5]{Center for Climate and Resilience Research (CR$^{2}$), Universidad de Chile, Santiago, Chile}

\address[6]{Departamento de Ciencias Forestales, Laboratorio de Ecología del Paisaje y Conservación, Universidad de La Frontera, Temuco, Chile}

\cortext[cor1]{Corresponding author}

\begin{abstract}
Monitoring wildfires is an essential step in minimizing their impact on the planet, understanding the many negative environmental, economic, and social consequences. Recent advances in remote sensing technology combined with the increasing application of artificial intelligence methods have improved real-time, high-resolution fire monitoring. This study explores two proposed approaches based on the U-Net model for automating and optimizing the burned-area mapping process. Denoted 128 and AllSizes (AS), they are trained on datasets with a different class balance by cropping input images to different sizes. They are then applied to Landsat imagery and time-series data from two fire-prone regions in Chile. The results obtained after enhancement of model performance by hyperparameter optimization demonstrate the effectiveness of both approaches. Tests based on 195 representative images of the study area show that increasing dataset balance using the AS model yields better performance. More specifically, AS exhibited a Dice Coefficient (DC) of 0.93, an Omission Error (OE) of 0.086, and a Commission Error (CE) of 0.045, while the 128 model achieved a DC of 0.86, an OE of 0.12, and a CE of 0.12. These findings should provide a basis for further development of scalable automatic burned-area mapping tools.
\end{abstract}


\maketitle
\section{Introduction}
\label{s:intro}
Monitoring wildfires is an essential step in efforts to understand and minimize their impact on the planet and the many negative environmental, economic, and social consequences of that impact \citep{masson2022global}. Both the risk of wildfires and their severity are expected to increase as a result of climate change, threatening natural and social values in various regions of the world \citep{@jiaipcc,moritz2012climate}. In a warming world, forested countries are particularly vulnerable to fire \citep{yuemd-8-1321-2015}. It has been estimated that 1.5°C of warming will increase fire emissions by 10.0\%-15.4\% \citep{tian2023projections}. Carbon neutrality accords such as the Paris Agreement and the recommendations of the Intergovernmental Panel on Climate Change (IPCC) are aimed at limiting the global average temperature increase over pre-industrial levels, but this goal is unlikely to be achieved given the low probability of success in current scenarios \citep{IPCC2022_last}. It is therefore urgent that climate change risks and impacts be reduced through the application of mitigation and prevention strategies, and this will require accurate monitoring of fires.

The technology for wildland fire monitoring has advanced significantly in recent years. In the past, satellite imagery was often used to detect and monitor fires, but the imagery was typically of low resolution with limited temporal and spectral coverage. Progress in remote sensing technology combined with the development of artificial intelligence has brought about a significant refinement of real-time, high-resolution, multispectral fire monitoring \citep{florath2022supervised,visser2004real}. Other technologies such as unmanned aerial vehicles (UAVs), ground-based sensors and even citizen science platforms are also being marshalled to provide more comprehensive and accurate wildfire data. Remote sensing in particular has contributed to the mapping of areas burned by wildfires \citep{yuemd-8-1321-2015}. This last technique has made it possible to assess the size and extent of burned areas as well as fire severity, thus providing insights into wildfires’ ecological and environmental impacts \citep{levin2020remote,nagendra2013remote, jaiml} and contributing, along with other technological advances, to early fire detection and response. The continued evolution of wildfire monitoring technology will undoubtedly play a critical role in managing the impacts of these events.

Burned-area global products typically incorporate traditional burned-area mapping models for determining fire scars that are built around rule-based algorithms and iterative workflows with coarse spatial resolution (>300 m). This has significantly limited their performance, as indicated by the high levels of commission errors (CE) and omission errors (OE) in their predictions \citep{padilla2015comparing}, in turn hampering their effectiveness in small-fire detection (< 100 ha) \citep{CHUVIECO201945}. Some traditional models have suffered similar performance limitations due to the use of low resolution imagery \citep{liu2018burned, roteta2019development}. More current models also using rule-based algorithms and iterative workflows but with medium spatial resolution still present difficulties in aspects such as the calibration of indices by different land-cover types \citep{ROY2019111254, Trisaktitech} or the identification of burned pixels when fire scars are patchy or pixel variability is particularly severe due to uneven fuel combustion, diverse vegetation or soil structures, or other factors \citep{CHUVIECO201945, liu2018burned}. These weaknesses result in misidentification of pixels, an inability to include an entire burned area in a single scar, and the failure to differentiate between wildfires and agricultural burnings \citep{boschetti2015modis, ROY2019111254}. 

Important advances relevant for burned-area mapping have recently been made with artificial intelligence. These include contextual analysis, which takes into account the pixel values of neighbouring pixels in a bigger picture instead of just categorizing pixels by spatial, spectral or temporal thresholds, and the use of previous information inputted to the model, thus allowing a more complete analysis \citep{alzubaidi2021review}.  

The development of machine learning (ML) and deep learning (DL) models has simplified iterative processes and improved their performance. Among the different types of DL models, convolutional neural networks (CNN) models stand out for their performance in applications such as image recognition, classification, segmentation, localization problems and feature extraction \citep{alzubaidi2021review}. ML and DL models have begun to be used in the last few years for burned-area mapping \citep{Huunirs13081509,jaiml} and other applications related to the understanding and management of wildfires \citep{carrasco2021exploring,carrasco2023firebreak,miranda2020evidence,pais2021deep}. CNNs have attracted attention for their high performance levels, using fewer data and less computation time \citep{Pintors13091608,Huunirs13081509}. U-Net model \citep{ronnenbergunet} implementations in particular \citep{debemrs12162576,Huunirs13081509, Knopprs12152422, shamsoshoara2021aerial} have performed extremely well compared to the previously mentioned models, none of which take contextual information of the same number of variables simultaneously into account. 

\cite{Huunirs13081509} compared different methods for burned-area mapping using a monotemporal approach (i.e., only post-fire images). Included in the comparisons were the DL models Fast-SCNN, DeepLabv3+, HRNet and U-Net, as well as 13 traditional ML model algorithms and others based on a spectral index. The tests revealed that HRNet, U-Net and Fast-SCNN performed best, due in general to their ability to take account of contextual information and in particular to their greater success in delineating fire scar perimeters, especially when the latter were clearly delineated from the outset. This superior performance is explained by the greater capability of DL models to analyze the relationships between a given pixel and its neighbours, differentiating between them on multiple criteria learned through model training. The neurons or kernels learn through the optimization function to identify when a particular attribute is relevant for a better prediction, and if it is not, a different weight will be assigned to it. Certain other attributes like position, shapes, sizes and colours can be taken into account in predicting a class or object. A number of weaknesses in the monotemporal approach were nevertheless also noted such as the failure to distinguish and differentiate agricultural burnings, a problem that has repeatedly been reported \citep{arruda2021alternative,Knopprs12152422}, and reburns (i.e., a new burn over the same area as a previous one). In the latter case, such distinctions are not possible using only post-fire data.  

In Chile, the focus of the present study, fires have had severe impacts over the last few decades, a situation clearly reflected in the upward trend in burned area (CONAF, 2022). \href{https://www.pangaea.de/tok/6dcc6e08241c5076ef6bff47bbe73014308d4881}{The Landscape Fire Scars Database for Chile} recently released information on more than 8,000 fire scars nationwide that was generated using a semi-automatic approach \citep{mirandaessd-14-3599-2022}. The availability of these data has created an excellent opportunity to assess methodologies for automating the mapping of burned areas. 

Taking advantage of this opportunity, the present paper develops an approach using the U-Net model that attempts to automate and optimize the burned-area mapping process. The model is fed by Landsat imagery and 1985-2018 time-series data for Valparaíso and BioBío, two prototype Chilean regions that have been seriously affected by fires in recent decades. To deepen our understanding, an additional objective of the present work is to devise a method that facilitates the future automation of the entire fire scar collection process from data collection to fire-scar prediction. With these goals in mind, two U-Net models reflecting variations on the proposed approach are presented and evaluated for two different data collection scenarios, one based on image sizes of $128\times 128$ pixels and the other on different sizes depending on the size of the fire scar. This difference between the models implies the study of different balances between the burned-area and unburned-area data classes, which reveals how much class balance influences the models' yield and contributes to further development of burned-area mapping tools. 

\section{Material and Methods}
\label{s:matandmet}
\subsection{Study area}
\label{ss:studyarea}

The raw dataset used with the U-Net models is taken from the Landsat imagery in \cite{mirandaessd-14-3599-2022}, which covers the 1985-2018 period for different land covers in ten of Chile’s sixteen administrative regions. As already noted, our area of study comprises the heavily affected Biobío and Valparaíso regions. The data contain Landsat pre- and post-fire images with a spatial resolution of 30 m and a temporal resolution of 16 days. 

\subsection{Remote sensing and fire scar data}
\label{ss:remotes}

The segmented fire scars corresponding to each model from \href{https://www.pangaea.de/tok/6dcc6e08241c5076ef6bff47bbe73014308d4881}{The Landscape Fire Scars Database for Chile} are in separate binary raster files containing the burned areas, with burned pixels assigned a value of 1 and unburned pixels a value of 0. Each pre- and post-fire image is composed of eight spectral bands: 1) blue, 2) green, 3) red, 4) NIR, 5) SWIR1, 6) SWIR2, 7) NDVI Eq.\:(\ref{eq:ndvi}), and 8) NBR Eq.\:(\ref{eq:nbr}). The first six bands are from the Landsat data while the latter two are spectral indexes calculated mathematically from other six. The fire scars constitute the reference data and are referred to as labels given that their pixels are labelled in the categories the model attempts to predict.

\begin{equation}
    NDVI=\frac{NIR-Red}{NIR+Red}
    \label{eq:ndvi}
\end{equation}

\begin{equation}
    NBR=\frac{NIR-SWIR2}{NIR+SWIR2}
    \label{eq:nbr}
\end{equation}

The Relative differenced Normalized Burn Ratio (RdNBR) index \citep{MILLER200766} is used to construct the fire scars \citep{mirandaessd-14-3599-2022} and determine their severity, and is given by
\begin{equation}
    RdNBR=\frac{NBR_{pre}-NBR_{post}}{\sqrt{\frac{|NBR_{pre}|}{1000}}}
    \label{eq:Rdnbr}
\end{equation}

\subsection{Data pre-processing and filtering}
\label{ss:datapp}

The methodology underlying the proposed approach is summarized by the flowchart in Fig. 1. It consists of three phases: pre-processing and filtering, preparation for the U-Net, and modelling. Each phase is in turn made up of various stages as shown in the figure.  

\begin{figure*}[ht]
    \centering
    \includegraphics[scale=0.8]{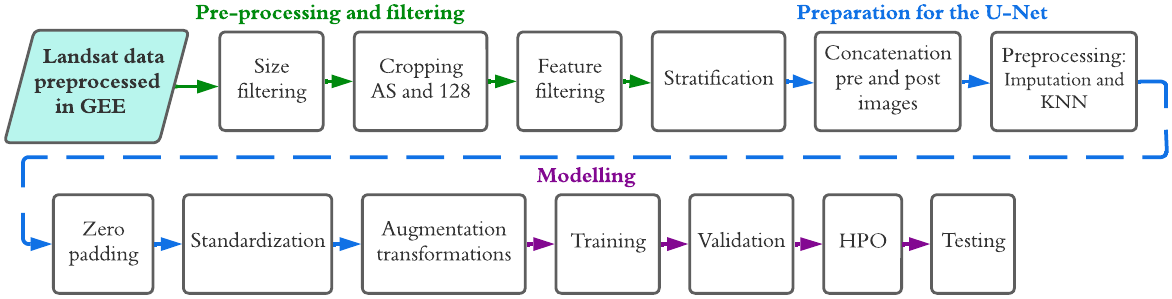}
    \caption{Methodology flowchart.} 
    \label{fig:diag}
\end{figure*}

\begin{figure*}[ht]
    \includegraphics[scale=0.60]{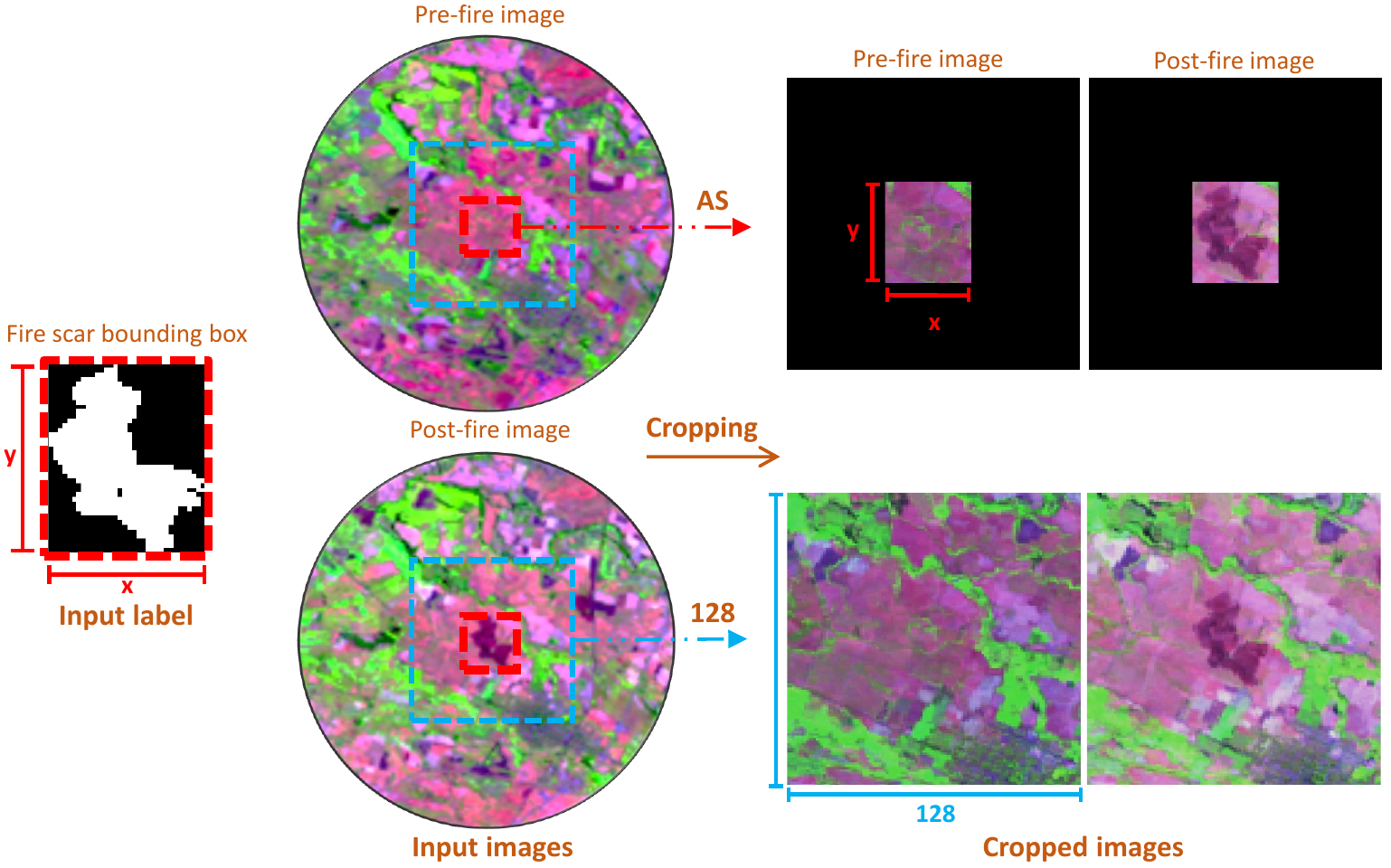}
    \caption{Cropping of images to create the AS and 128 datasets. The blue dashed square marks the 128 dataset borders; the red dashed square marks the AS dataset borders that limit the tile to the fire-scar bounding box.}
    \label{fig:Data}
\end{figure*}

The two variations on the proposed approach are compared in terms of the models’ performance under two different data gathering and pre-processing scenarios (Fig.\:\ref{fig:diag}). In each scenario, a different image size is inputted depending on the bounding box of the existing fire scars, thus resulting in a different class balance and therefore a different burned proportion in each image. The two corresponding datasets were created using the original pre- and post-fire images that were then cropped to different dimensions  (Fig.\:\ref{fig:Data}). The first dataset was cropped to the bounding box area of the fire-scar rasters (i.e., the quadrilaterally shaped fire-scar raster boundaries) and denoted AllSizes (AS), while the second one, named simply 128, was cropped to the standard $128\times 128$ pixel size in order to facilitate the subsequent filtering process and the managing and processing of the data. Only fire scars with fewer than 128 pixels on either of the $x$ and $y$-axes, which accounted for the majority of the available files (91\%), were selected. The definitive tiles were obtained by extending them from the bounding box along each axis to 128 pixels. For both datasets the extension was evenly distributed, placing the fire scar raster at the centre and then extending the tiles left and right along the horizontal axis, and up and down along the vertical axis. The AS dataset thus obtained included an average burned area in terms of pixels of 32.9\% while for the 128 dataset the figure was 3.4\%, the latter case thus being substantially more imbalanced in terms of burned and unburned pixel classes.

Comparisons of the pre- and post-fire images, which were separated by a time gap of at least 16 days as determined by the satellite’s temporal resolution, revealed a number of recurrent features including reburns, multiple fires from different dates in a single image, and agricultural burnings. A feature that filters the data was applied for this purpose. The model automatically recognizes fire-scar areas using satellite images processed with Google Earth Engine (GEE) obtained through the precise GPS coordinates of each fire event. 

To facilitate post-processing, the model was trained to identify reburns. This was accomplished using a multitemporal approach that compared pre-fire images with post-fire ones. Also, since it was desired to study each fire scar individually, the model was trained to recognize a single fire per image located approximately at its centre. In this regard, a problem that frequently arises due to Landsat’s 16-day temporal resolution is that the satellite images contain multiple fire scars far enough apart to suggest that they were caused by independent events. However, given that additional burned areas, though not directly attached either to the central one or to one another may in fact be the result of the same event, it was decided that such additional areas would be considered as independent and filtered out only if they were more than 500 m from the central one.

The data were also checked for the presence of agricultural burnings. A non-automatic visual inspection of every image was conducted to filter out those where the corresponding labels included such burnings as part of the fire scar areas. The method for distinguishing them in an area containing fire scars consisted in comparing the shape of the burned areas and their uniformity in the pre- and post-fire images. 

Upon completion of the filtering process, the raw dataset contained a representative distribution of 977 fires in Valparaiso and 989 fires in Biobío, for a total of 1,966.

A dataframe was designed to order and distribute the inputting of the data so that each batch entering the model was representative of the dataset. Since the size of the burned areas was highly variable it was chosen as the stratification factor, thus ensuring each batch contained scars from the smallest as well as the largest size ranges. The data stratification was executed separately for the two regions in the study, thus creating a subdataset for each one. 
 
\subsection{Data preparation for the U-Net}
\label{ss:datap}
The pixel values of the images were inputted to the models in 16 concatenated array tensors (Fig.\:\ref{fig:diag}, Concatenation pre and post images), the first eight of which were for the post-fire image spectral bands while the second eight were for the pre-fire ones (Section \ref{s:matandmet}). The initial preprocessing of the data (Fig.\:\ref{fig:diag}, Preprocessing: Imputation and KNN) was required due to the presence of values outside the valid range for the Landsat images \citep{usgs57,usgs8}, anomalous values or outliers, invalid or not-a-number values (NaNs), and noise (appearing as zeroes). In the case of the off-range values and outliers, values were imputted as averages of each band for each image. Limits were calculated from the values of each band in the datasets based on the distribution of the complete dataset values, and values located at least 1.5 times the interquartile range of the average values were replaced by the average values of the corresponding band in each image. For the NaNs and noise, the k-nearest neighbours algorithm (KNN) was used to improve visualization. 

To be inputted to the models, the images had to have a fixed input size, which was set to $128\times 128$ pixels. While the pixel values within the fire-scar bounding boxes were the same for both the AS and 128 datasets, the pixel values outside of them were not, and for AS, these pixels were filled out with zeroes as the value (Fig.\:\ref{fig:diag}, Zero padding). Since the two datasets have different characteristics, they were processed independently in order to be inputted to the corresponding models and trained correctly. Two normalization methods, standardization and min-max scaling, were tested. This was done separately for each model given their different pixel values since those for pixels outside the fire-scar bounding boxes were included. In the case of dataset 128, they are the ``context'' values. The tests found that standardization performed best and was thus the method chosen (Fig.\:\ref{fig:diag}, Standardization). 

In situations such as the one studied here where data are limited, augmentation is widely applied in DL models to increase the amount of model learning. It involves techniques such as modifying the existing data by applying geometric transformations or altering the colours or brightness level, thus creating duplicates that are slightly different so the model can continue learning from the data. This improves generalization and reduces overfitting \citep{perez2017effectiveness}. In the present case, the data were subjected to flipping and rotation transformations (Fig.\:\ref{fig:diag}, Augmentation transformations).  

\subsection{Modelling}
\label{ss:u-net}
The modelling phase, in which the images are semantically segmented by classifying the pixels based on their category labels \citep{howdeeplearning}, consists of four stages:
training, validation, hyper-parameter optimization (HPO), and testing. In the training stage, the model learns the attributes of a fire, defining the parameters that represent the interactions between the image pixel values of the training set. Following this is the validation stage, in which the model evaluates its performance using the parameters learned during the training stage (Fig.\:\ref{fig:diag}, Training and validation). This involves minimizing the loss, calculated on the validation dataset, between the reference values, in this case the real values, and the above-mentioned pixel values that the model is attempting to predict automatically. The calculation of the parameters is reiterated until the minimum loss is obtained for the training and validation period. Then, in the HPO stage (Fig.\:\ref{fig:diag}, HPO), the hyper-parameter is tuned until the best results are obtained. Lastly, the model is tested using an independent dataset to calculate the model performance metrics (Fig.\:\ref{fig:diag}, Testing).

The distribution of the data from each of the AS and 128 datasets across the training, validation and testing stages was 70/20/10\%, respectively. The model for each dataset was then trained, validated and tested separately using the same data records but with the images cropped to each one's corresponding size. The two models were designed to test the effect of class balance on the performance of the proposed methodology and to determine their suitability for a future workflow that would include the integration of the satellite data using the available satellite coordinates such that the fire scar is relatively well-centred in the image tiles.

\begin{figure*}[!ht]
    \includegraphics[scale=0.7]{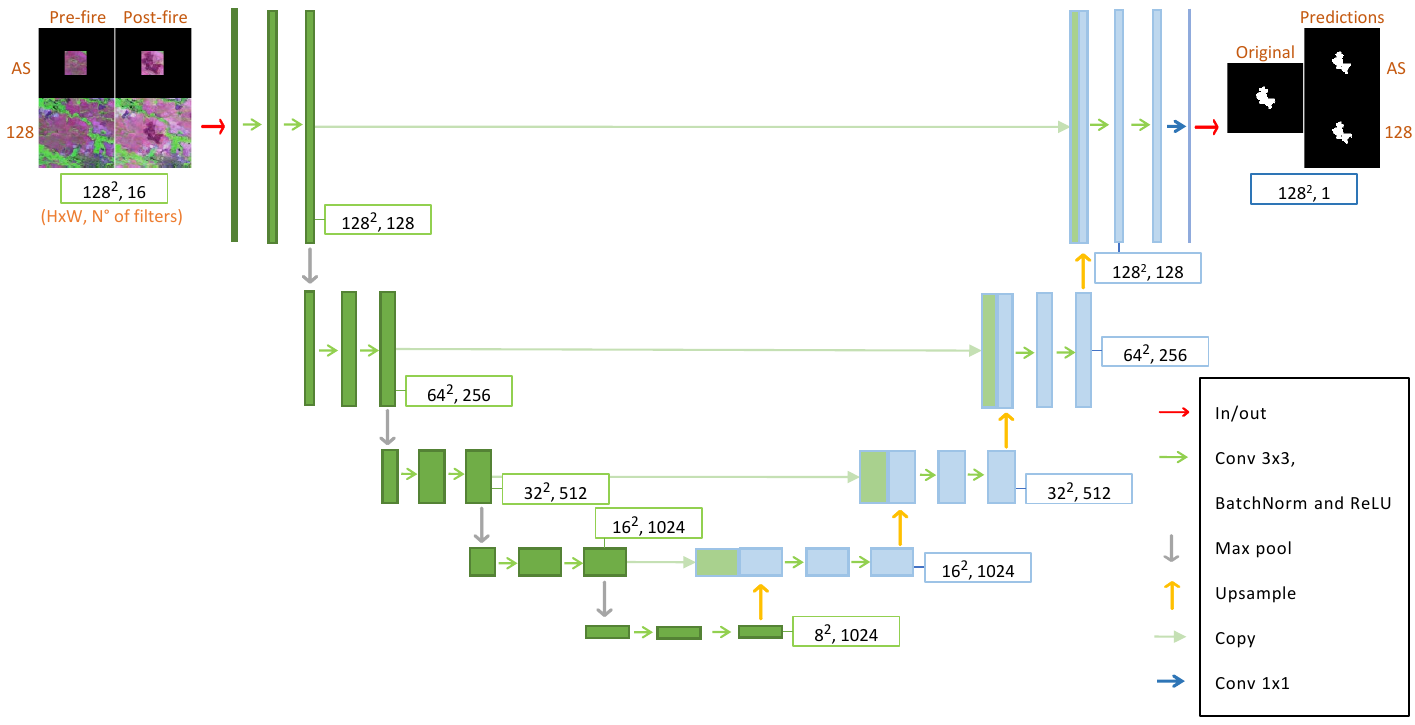}
\caption{Proposed U-Net architecture. The various operations are identified by the coloured arrows defined in the legend. The procedure for the AS and 128 models is the same except for the input and output data sizes, which are shown in the images (AS above, 128 below).}
    \label{fig:unet}
\end{figure*}

Iterations were run in order to determine all of the hyperparameters except the loss function and the number of epochs, which were decided arbitrarily based on standard CNN model applications. A preliminary configuration of hyperparameter values was obtained that served as the baseline for subsequent testing with a view to their optimization. These initial values were as follows: a learning rate of 1e-4, a batch size of 16, a total of 128 initial filters (which number doubles at every block of the encoding pathway except the last one on the downside, and halves at every block on the upside); an augmentation factor of 2, and 25 epochs. The optimizer was adaptive moment estimation (Adam) \citep{kingma2014adam} and the loss function was binary cross entropy (BCE). The optimization of the hyperparameters was achieved by iterative testing of them to improve model performance based on the Validation DC (Val DC) metric (Section \ref{ss:metrics}). The U-Net architecture is depicted in Fig.\:\ref{fig:unet}. The input size was set at $128\times 128$, which meant that for the AS model, a zero-padding function was applied for each array in order to fill out the dimensions.

The code developed for the models, including the pre-processing and filtering phase and the training, validation and testing stages, was initially based on \cite{mommertind} using Pytorch \citep{paszke2019pytorch} as a framework, and was then modified. The least-modified sections of the code were those relating to the pre-processing and filtering phase, the concatenation, standardization and augmentation stages in the preparation for U-Net phase, and the training and validation stages in the modelling phase. The preprocessing (imputation and KNN) and zero padding sections were redesigned and the testing section was significantly modified to suit the present application, considering factors such as the nature of the data, the desired metrics and improvement of visualization. The U-Net architecture differed from the original in the use of zero padding, $2\times 2$ upsampling through bilinear interpolations in the decoding pathway, and the exclusion of the middle hidden filters in the deepest convolutional block while maintaining its number of filters. 

The models were run on a computer featuring an AMD Ryzen 9 5900X 3401 MHz processor with 16 main processors, 32 logic processors and 32 GB of RAM. The operating system was Windows 10 and the graphics card was an NVIDIA GeoForce RTX 3080 Ti.

\subsection{Testing metrics}
\label{ss:metrics}
The metrics typically used to test model results are based on the numbers of pixels correctly and incorrectly classified in each category (e.g., burned or unburned) as determined by comparisons with the reference values. The comparisons label each pixel as a true positive (TP), true negative (TN), false positive (FP) or false negative (FN). The formulae for the metrics are given in Eqs. (\ref{eq:ce}), (\ref{eq:oe}) and (\ref{eq:dc}).

\noindent Commission error ratio: 
\begin{equation}
    CE=\frac{FP}{TP+FN}
    \label{eq:ce}
\end{equation}
\noindent  Omission error ratio:
\begin{equation}
    OE=\frac{FN}{TP+FN}
    \label{eq:oe}
\end{equation}
\noindent Dice Coefficient (DC):
\begin{equation}
    DC=\frac{2TP}{2TP+FP+FN}
    \label{eq:dc}
\end{equation}

The DC is a statistical measure that combines the OE and CE, and is a widely used metric in remote sensing applications. It serves as an effective tool for evaluating the accuracy and similarity of image segmentation results, especially in scenarios where precise delineation of objects or regions of interest is critical.

Finally, the training loss and validation loss metrics were used for evaluating the training and validation stages. They were calculated for each epoch using the BCE on the corresponding stage, with values ranging from 0 for the worst performance to 1 for the best. The minimal loss is the smallest possible error as indicated by the comparisons of the validation predictions with the reference data or ground truth. The validation loss was used to select the best model in each stage.

\subsection{Hyperparameter optimization and configuration}

Hyperparameter optimization is an essential step in obtaining good model performance. To determine whether the baseline configuration (Section \ref{ss:u-net}) was optimal, three tests were conducted on three hyperparameters: number of filters or layers in the network, batch size, and learning rate. The configurations were evaluated using the Val DC metric and the results are shown in Table \ref{tab:hpo}. As can be seen, the baseline configuration performed best and thus was chosen for the model.

\begin{table*}[ht]
    \centering
    \caption{HPO Tests}
    \begin{tabular}{@{}cccccccc@{}}
    \toprule
 Name                                                    & Configuration              & Val DC$^a$                        \\ \midrule
    Baseline performance                                    & Base HPO$^b$                   & 0.948  \\ \midrule
    Filters$^b$ & 32/256                     & 0.937        \\                 
           & 160/1280                   & 0.946
           \\ \midrule
         Learning rate & 1e-5                & 0.92          \\
           & 1e-3                       & 0.942
           \\ \midrule
         Batch size                                          & 10                         & 0.947            \\            
            & 24 & 0.942

\\ \bottomrule
    \label{tab:hpo}
    \end{tabular}
\\
\footnotesize{$^a$ Val DC is the DC value calculated for the predictions of the validation dataset images. $^b$ The base HPO used the following configuration: a learning rate of 1e-4, a batch size of 16, 25 epochs, 128 initial filters and an augmentation factor of 2. $^c$ In the filter configuration x/y, the x is the number of initial filters in the first convolutional layer and the y is the number in the deepest layer.} 
    \\
\end{table*}

As regards the filters, the more there are in each model layer, the more parameters there are, and therefore the greater is the impact on computation time. In the present case, however, performance improved by 1.1\% when the number of filters was increased from 32 to 128, the quantity which gave the best result. It was therefore used for the first layer, duplicating the number of filters through to the last encoding pathway layer, the deepest one in the network. The learning rate was the hyperparameter with the greatest impact on Val DC due to the fact that the weight adjustment factor in the Adam optimization function is highly dependent on the hyperparameter itself. Finally, although batch size was not found to be sensitive in this model, the variable does impact memory use. A value of 16 was chosen.

\section{Results}
\subsection{Training, validation and hyperparameter optimization }

\begin{figure*}[ht]
    \centering
    \includegraphics[scale=.65]{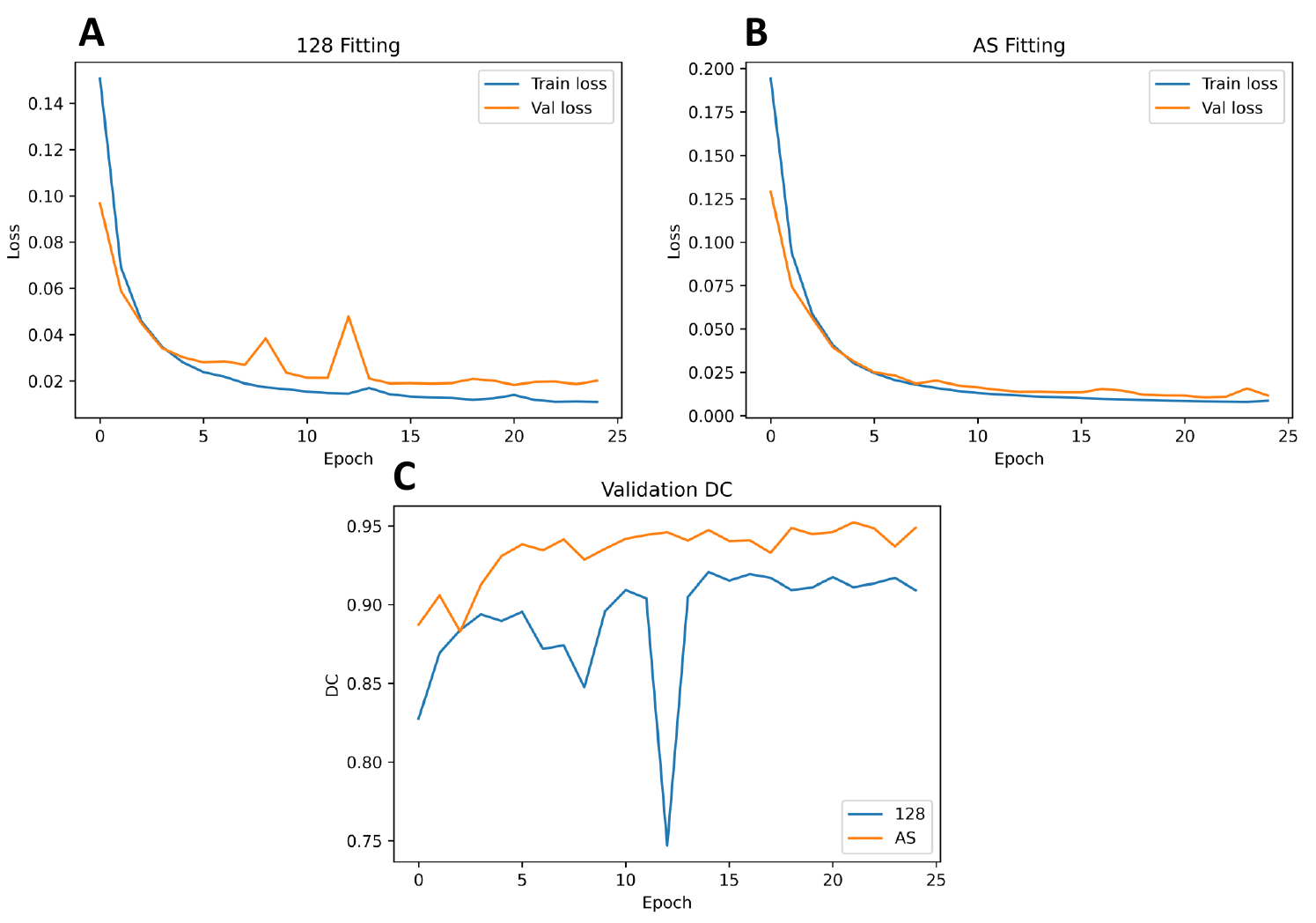}
    \caption{Training and validation curves. Charts \textbf{A} and \textbf{B} plot the models’ fit, showing the loss values during the validation and training stages. \textbf{C} shows the model’s validation DC by epoch.}
    \label{fig:loss}
\end{figure*}

The performance results of the AS and 128 models using the best hyperparameter configuration found by HPO and an augmentation factor of 3 are shown in Table \ref{tab:summres}. Initially, AS performed better than 128 on all metrics. The validation DC for the two models, plotted in Fig.\:\ref{fig:loss}-\textbf{C}), indicates the stochasticity associated with them and the trend in the learning process as training increases using the Adam optimizer \citep{kingma2014adam}. Beginning around epoch 13, little additional improvement is registered. Both models got better results on CE than OE. As regards training and validation, the curves were neither overfitted nor underfitted (Fig.\:\ref{fig:loss}-\textbf{A} and \textbf{B}).

\begin{table*}[]
    \centering
    \caption{Summary of training, validation and testing results for the AS and 128 models.}
\begin{tabular}{@{}llll@{}}
\toprule
\multirow{4}{*}{} &       & AS         & 128          \\
\midrule
\multirow{4}{*}{Training and validation} & Epoch      & 22          & 15          \\
                                         & Train loss$^a$ & 0.008       & 0.014       \\
                                         & Val loss$^b$  & 0.011       & 0.019       \\
                                         & Val DC     & 0.952       & 0.921       \\
\midrule
\multirow{4}{*}{Testing}                 & DC         & 0.930       & 0.863       \\
                                         & CE         & 0.0858      & 0.146       \\
                                         & OE         & 0.0446      & 0.0920     
\\ \bottomrule
    \label{tab:summres}
\end{tabular}
\end{table*}

The distributions by DC values of the average pixel values in each spectral band of the 195 test images for the fire-scar areas reference data (i.e., only burned pixels) were studied in order to compare the pre-fire/post-fire changes in those areas as predicted by the two models. The idea was to determine the impact of certain fire attributes on the models’ performance. The distributions are presented in Figs.\:\ref{fig:bandas1} y \ref{fig:bandas2}. As can be seen, the bands showing the greatest variation between the pre- and post-fire images were NIR, SWIR2, NDVI and NBR. Note in particular that the performance of the 128 model improved when the distance between values increased, showing greater sensitivity than the AS model. As regards the variation in average band values between the high and low performance quartiles, in the post-fire images the Blue, SWIR2 and NBR bands displayed particularly high levels, while in the pre-fire images SWIR2 showed the most variation in both models.  

Finally, Fig.\:\ref{fig:otros} presents a more detailed view of the test tiles performance in terms of DC plotted against burned area for both models and against total area for AS only, total area being a variable in the latter case as opposed to 128 where it is a constant at $128\times 128$ for all images. The three plots reveal that performance tended to improve as class balance equalizes or as the proportion of burned versus unburned pixels increased.

\subsection{Testing}
\begin{figure*}[ht]
    \centering
    \includegraphics[scale=1]{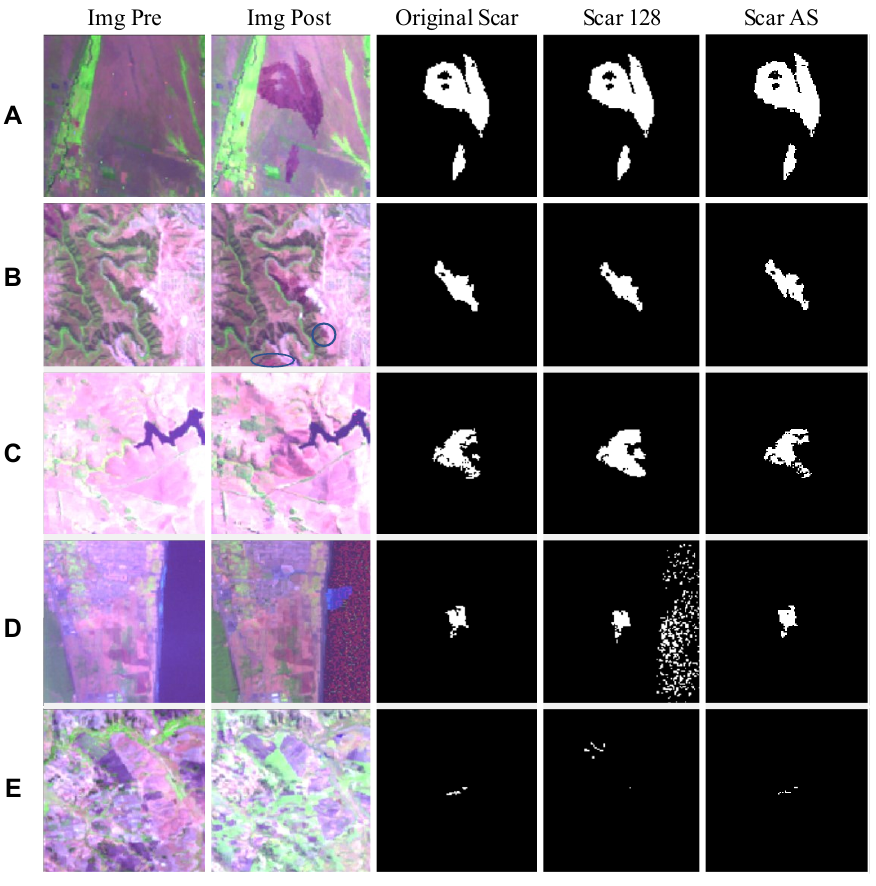}
    \caption{Test performance. The tiles shown are representative of the two models' best, average and worst predictions. The blue circles in the post-fire images in \textbf{B} (``Img Post'') mark areas not included either in the reference areas or in the predictions.}
    \label{fig:testingpe}
\end{figure*}

A number of representative predictions of the two models are displayed in Fig.\:\ref{fig:testingpe}, with the corresponding metrics set out in Table \ref{tab:testingtil}. One of the best performance levels of either model was exhibited by the tile in Fig.\:\ref{fig:testingpe}-\textbf{A}, which had a DC of close to 1 and a low CE value. Positive values of OE were predicted in almost all other cases, especially where visually observable burn severity was variable. 

In general terms, AS was more conservative than 128 with lower CE, detecting more precisely the unburned islands or spaces within the fire scar perimeters. There were, however, two cases where AS performed worse, as may be seen in Figs.\:\ref{fig:testingpe}-\textbf{B} and \textbf{C} where its OE values are lower.

In a number of instances, 128 predicted other burned areas located more than 500 m from the boundary of the main fire as part of the scar, which was undesirable in terms of the data selection criteria (Section \ref{s:matandmet}). By contrast, AS rarely predicted such areas, identifying in almost every image the one fire included in the reference data and no others. This was manifested in the major differences between the two models’ respective performance levels (see  Figs.\:\ref{fig:testingpe}-\textbf{D} and \textbf{E}). In a few cases, however, 128 did discern and predict only the principal fire (Fig.\:\ref{fig:testingpe}-\textbf{B}) and successfully excluded areas more than 500 m from its boundary, thus allowing segmentation of a single main fire per input centred in relation to the tile’s geographical coordinates.

Note finally that the worst performances appear in Figs.\:\ref{fig:testingpe}-\textbf{D} and \textbf{E}. In the former case, 128 mistakenly identified pixels representing water as burned. It occurred only in this test dataset tile and may be attributed in part to a directly proportional relationship in the pre- and post-fire data comparisons between the water and burned-area pixel values in the NIR and RED bands. As for the latter case, 128 managed to detect little of the burned area while AS had a relatively low DC of 0.69.

\begin{table}[]
    \centering
    \caption{Test results metrics. The rows are ordered numerically by the model 128 DC values from highest to lowest.}
    \label{tab:testingtil}
\begin{tabular}{@{}lllll@{}}
\toprule
                   &    & 128  & AS   &  \\ \midrule
\multirow{3}{*}{1} & DC & 0.99 & 0.99 &  \\
                   & OE & 0    & 0    &  \\
                   & CE & 0.02 & 0.03 &  \\
                   \midrule
\multirow{3}{*}{2} & DC & 0.93 & 0.91 &  \\
                   & OE & 0.13 & 0.16 &  \\
                   & CE & 0    & 0    &  \\
                  \midrule
\multirow{3}{*}{3} & DC & 0.86 & 0.90 &  \\
                   & OE & 0.03 & 0.17 &  \\
                   & CE & 0.22 & 0    &  \\
                 \midrule  
\multirow{3}{*}{4} & DC & 0.27 & 0.91 &  \\
                   & OE & 0.23 & 0.13 &  \\
                   & CE & 0.84 & 0.04 &  \\
                  \midrule
                   & DC & 0.04 & 0.69 &  \\
5                  & OE & 0.96 & 0.48 &  \\
                   & CE & 0.96 & 0    &  \\ \bottomrule
                   
\end{tabular}
\end{table}

\section{Discussion and conclusions}

As was explained in Section 2, the two image datasets were constructed and filtered with the objective of predicting only areas burned by wildfires, as opposed to agricultural burnings, and only the main fires, excluding any located at a distance from them of more than 500 m. This latter criterion was adopted to facilitate the post-segmentation process in cases where there were multiple fires from different dates in the same image in order that the analysis could focus exclusively on the main one. The occurrence of erroneous predictions on both objectives underlines the importance of applying well-defined criteria for the burned area it is desired to predict and constructing large and high-quality datasets for DL model training that are representative of the fire regime in the area under study. Under such conditions, the models will be able to make predictions for any size and severity range. Of the two models studied, 128 obtained better scores than AS on the metrics when the criterion of identifying a single main fire based on distance considerations was not applied or when use was made of large amounts of data, including agricultural fires outside of the fire scars and not counted as part of the reference data. 

In broad terms, both models scored better on CE than OE, which may be explained in part by the use of criteria for obtaining the reference data that favoured errors of omission over those of commission. Another general cause of errors, observable in Fig.\:\ref{fig:testingpe}-\textbf{A}, can be attributed to the input images, which were obtained through the application of different criteria employed to define the variables for recognizing burned pixels (based on the RdNBR threshold and other hyperparameters \citep{mirandaessd-14-3599-2022}). This reflects the difficulty in establishing very precise criteria for obtaining fire scars, particularly given that it is an iterative process which depends on the user’s definition of such scars based on visual observation. It further points up one of the more notable limitations of these models, which is their inability to make an accurate segmentation within the fire area perimeter where there are unburned islands not detected as such and therefore included among the burned pixels, thus resulting in commission errors.

The results also indicated that the average pixel value in the NIR, SWIR2, NDVI and NBR bands varied significantly between the pre- and post-fire images. A tendency to achieve better scores when these average value variations were relatively great was also observed, especially with 128. This finding validates the use of these bands and the impact of certain fire characteristics on this model. Another aspect evaluated was the variation in the average pixel value when comparing the high and low performance quartiles. The Blue, SWIR2 and NBR bands displayed particular high levels, while in the pre-fire images, SWIR2 showed the most variation in both the AS and 128 models. This implies that with the use of these bands, relatively severe fires will be detected more easily than less severe ones.

As already noted, the accuracy in identifying unburned islands was one of the limitations of the models. It was reflected in the lack of consistency in the testing-stage predictions and is directly attributable to the training data and the RdNBR threshold used in obtaining these data to determine the burned pixels based on the user’s visual observation. This is of particular concern given the high ecological values in the zones under study \citep{romanarticle}, and points up the need for further research using other burned-area mapping models. 

As regards the misidentification of pixels in areas covered by water (Fig.\:\ref{fig:testingpe}-\textbf{D}), this has been observed in other burned-area models such as \cite{Knopprs12152422}. But in the present case it was likely due to the directly proportional relationship in the NIR and RED bands between the burned-area and water pixels. Finally, in the case of Fig.\:\ref{fig:testingpe}-\textbf{E}, the errors might have been produced by the low severity of the fire as indicated by the RdNBR value of 200, the small size of the burned area, and the possibility that it was in fact not a fire but rather an agricultural burning.

\subsection{Further development of proposed models}

Further development of the two fire-mapping methodologies presented here could strengthen their ability to include the mapping of burned areas based only on the geographical coordinates of a fire’s centre. Depending on how precise are the coordinates, the 128 model could potentially map fires extending up to 3.84 km along each axis using satellite data, with burned areas located at the centre of the image. Alternatively, another image preprocessing function could be incorporated into the AS model that would permit the inputting of satellite images which are larger and therefore contain more context, with an automatic feature to create bounding boxes for the burned areas. Artificial intelligence models with similar abilities have been developed in applications aimed at the detection of objects in satellite images (see, e.g., \cite{qianarticle},\cite{yoloSharma2021YOLOrsOD} or \cite{martins2022deep}). The improvement in performance attained by an AS model with these capabilities is potentially greater than that achieved by simply inputting additional contextual area, as was observed in the results presented here.

The use of Sentinel-2 data is also recommended as a way of improving model performance given that it offers a spatial resolution of 10 m. The transferability to satellite data of DL models, by which is meant models trained with data from a satellite that are then used to make predictions using data from another satellite, has been demonstrated in \cite{Huunirs13081509} and \cite{martins2022deep}. In the first study, the authors were able to transfer a U-Net model trained with Sentinel-2 data to the corresponding Landsat data while maintaining metrics within acceptable ranges ($<5\% $ variation in Kappa). In the second study, a novel approach was proposed to map burned areas using high-resolution PlanetScope imagery and knowledge transfer from Landsat-8 imagery. The results showed a significant improvement in the accuracy of burned-area detection compared to traditional methods, demonstrating the potential of deep learning techniques and knowledge transfer in the field of remote sensing, as well as a potential application of the models proposed here.

For fires extending more than 3.84 km along each axis, the distribution of the geographical coordinates could be carried out previous to data processing in order to generate more tiles that could then be combined. Another recommendation would be to implement post-processing tools in order to improve information at the boundaries of the predictions, check pixel consistency, etc., as shown in \cite{cheng2020novel}.

In conclusion, this study demonstrated that an automatic approach can optimize the process of burned-area mapping with a low error rate through the application of a direct methodology that uses Landsat images displaying fire scars. The more balanced is the input data to the proposed modelling, the better will be the predictions. Given the variety of landscapes tested and the generally good performance obtained, we may assume that the approach could be successfully applied to data for fires of a size not less than the average analyzed in this study and originating in locations with similar landscapes to those found in the Chilean regions of Valparaiso and Biobío, used here as test cases. 

The results also showed that obtaining a good performance level in burned-area mapping requires data on the fire events’ geographical coordinates, which if available in a database would permit the automation of the entire process from data collection through to prediction. Thus, the methodology presented here could serve as a skeleton for the development of more advanced approaches, including data collection with deeper pre-processing to locate the fire scar at the centre of the image and the tiling process previous to prediction.

\printcredits

\section*{Declaration of competing interest}
The authors declare that they have no known competing financial interests or personal relationships that could have appeared to influence the work reported in this paper.

\section*{Acknowledgments}

This project has received funding from the European Union’s Horizon 2020 research and innovation programme under grant agreement No 101037419 (\href{https://fire-res.eu}{FIRE-RES}). The authors acknowledge the support of the Agencia Nacional de Investigaci\'on y Desarrollo (ANID), Chile, through project FONDEF  ID20I10137.
\textbf{JC} acknowledges the support of the ANID, through funding Postdoctoral Fondecyt project No 3210311.  \textbf{AM} acknowledge
funding from ANID/FONDAP (grant No 15110009) and to the ANID postdoctoral Fondecyt project (grant No 3210101). Special thanks are extended to Kenneth Rivkin for his invaluable assistance in preparing the English translation of this manuscript. His expertise and attention to detail greatly improved the clarity and quality of the final version.

\section*{Data and code availability}
The datasets and Python code used to reproduce our analyses can be accessed at the following GitHub repository: \url{https://github.com/fire2a/FireScars}.

\bibliographystyle{cas-model2-names}
\bibliography{cas-refs}

\clearpage

\section*{Supplementary material}
\begin{figure*}[h!]
    \centering
    \includegraphics[scale=.93]{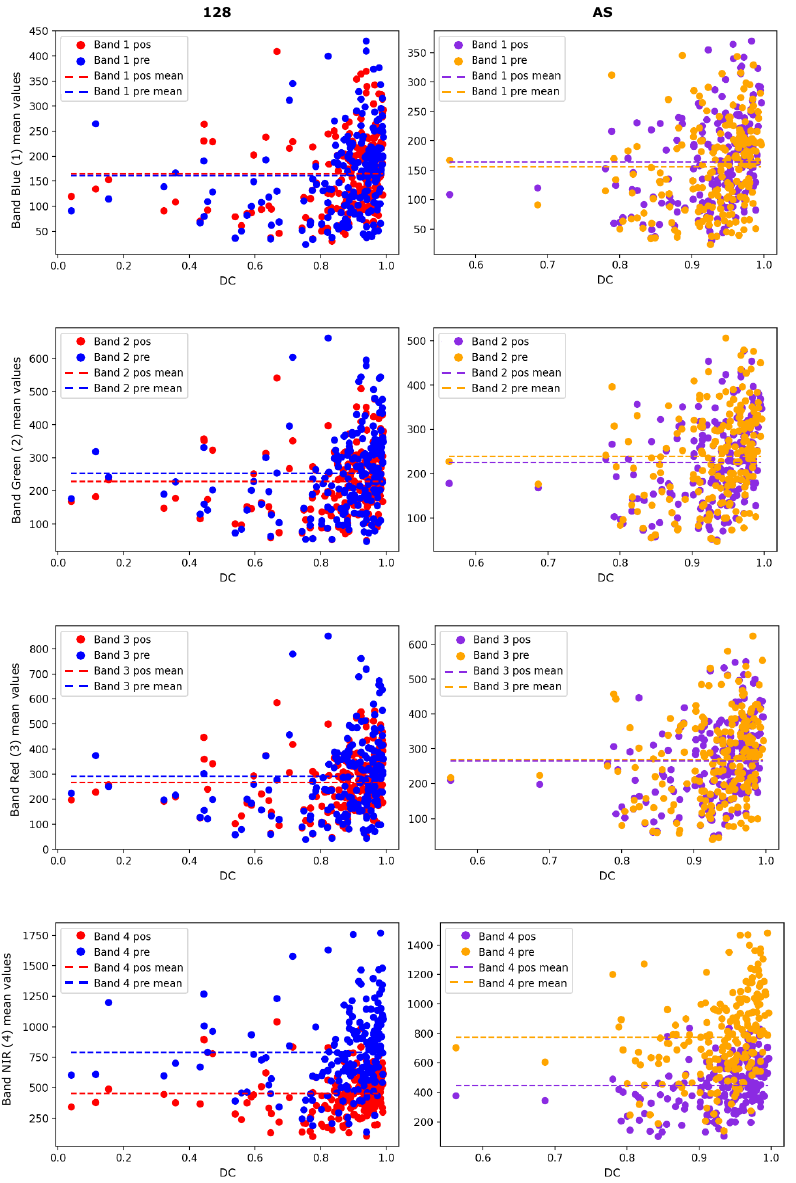}
    \caption{Average values of the testing images vs. performance (DC), Bands 1-4.}
    \label{fig:bandas1}
\end{figure*}

\begin{figure*}[h!]
    \centering
    \includegraphics[scale=.93]{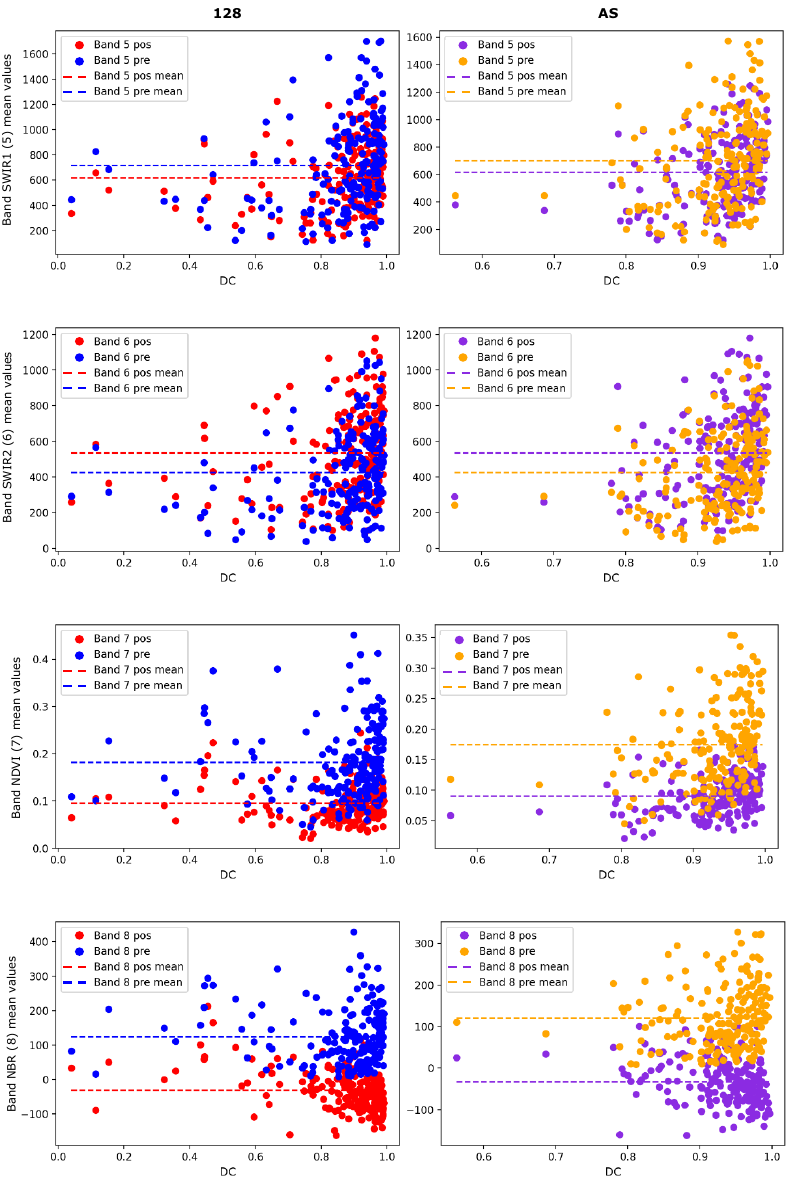}
    \caption{Average values of the testing images vs. performance (DC), Bands 5-8.}
    \label{fig:bandas2}
\end{figure*}

\begin{figure*}[h!]
    \centering
    \includegraphics[scale=0.95]{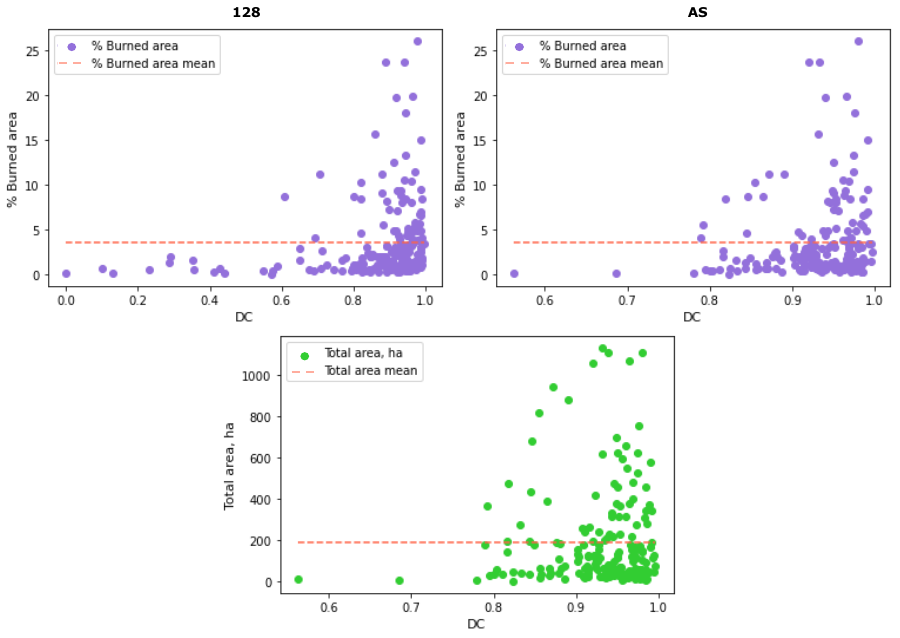}
    \caption{Test performance vs. burned area and total area.}
    \label{fig:otros}
\end{figure*}

Figures \ref{fig:bandas1} and \ref{fig:bandas2} show the distribution of the average values of each band in the test images vs. the DC obtained in the predictions. Figure \ref{fig:otros} shows the \% burned area values and (for the AS model only) the total area in the images vs. the DC. For all three figures, the outliers of the average band values were replaced by the mean values of the corresponding band for better visual clarity.

\end{document}